# POP-CNN: Predicting Odor's Pleasantness with Convolutional Neural Network

Danli Wu[1],
wdanli2018@outlook.com

Yu Cheng[1*],
chengyu@gdut.edu.cn

Dehan Luo[1],
dehanluo@gdut.edu.cn

Kin-Yeung Wong[2],
akywong@ouhk.edu.hk

Kevin Hung[2],
khung@ouhk.edu.hk

Zhijing Yang[1]
yzhj@gdut.edu.cn

1 School of Information Engineering, Guangdong University of Technology

2 School of Science and Technology, The Open University of Hong Kong

**Abstract**—Predicting odor's pleasantness simplifies the evaluation of odors and has the potential to be applied in perfumes and environmental monitoring industry. Classical algorithms for predicting odor's pleasantness generally use a manual feature extractor and an independent classifier. Manual designing a good feature extractor depend on expert knowledge and experience is the key to the accuracy of the algorithms. In order to circumvent this difficulty, we proposed a model for predicting odor's pleasantness by using convolutional neural network. In our model, the convolutional neural layers replace manual feature extractor and show better performance. The experiments show that the correlation between our model and human is over 90% on pleasantness rating. And our model has 99.9% accuracy in distinguishing between absolutely pleasant or unpleasant odors.

## Ⅰ. INTRODUCTION

As we all know, the smell has a pivotal position in human life, but there has been a lack of appropriate words to describe odor, Plato thinks "the varieties of smell have no name, but they are distinguished only as painful and pleasant" in the Timaeus [1]. That is to say, the basic phenomenological object of olfaction is not something "what is", it is a kind of perception [2]. The perception is closely related to our natural emotions and acquired learning. Some people or animals are naturally sensitive to certain odors, Dielenberg's laboratory mice can prove this [3]. And perception is largely plastic, dependent on acquired experience and learning, and influenced by culture, emotion and even gender [4] [5]. Olfactory researchers are working to find out the relationship between odor's stimulation and perception, which has attracted much attention because pleasantness is the main axis of perception. By studying whether odors give people a feeling of "pleasant" or "unpleasant", and the magnitude of the degree, establishing predictive models of pleasantness can simplify the evaluation of new odors. In addition, these models have practical application functions. When perfuming, it can reduce the subjectivity brought by different people's preference for a certain type of fragrance, and save time or manpower. In environmental monitoring, there is data to prove that the hazard of odor is related to its pleasantness to some extent.

The exploration of odor's pleasantness mainly focuses on the physicochemical characteristics of gases, and there are few studies on predicting the pleasantness using E-nose. Rehan M. Khan et al. studied the prediction of odor's pleasantness from molecular structure in 2007 [6], which is the pioneer of predicting odor's pleasantness. They used multiple PCA to find the correlation between molecular space and linguistic space, which has relatively large limitations. Rafi et al. wanted to obtain odor pleasantness from the angle of the E-nose in 2010 [7], which is completely different from other related odor research, and provides new ideas and methods for odor perception research. But he used the manual feature extraction method, which. requires the extraction algorithm and has poor versatility. In 2014, Ewelina Wnuk used the exemplar listing, similarity judgment and off-line rating to confirm that the odor term in Maniq (Maniq is a language spoken by a few nomadic hunters in southern Thailand.) has the complex meaning of coding odors, and these terms are coherent and the underlying dimensions are pleasantness and dangerousness [8]. As Plato said, "painful and pleasant." In 2016, Andreas Kelle et al. proposed a very powerful psychophysical dataset and used it to link the physicochemical characteristics with olfactory perception, and found that humans have a certain degree of correlation with the familiarity of odors and the description of perception. The familiar odor depends on the previous memory, and the unfamiliar odors are generally rated as neither "unpleasant" nor "pleasant"[9]. Subsequently, Kobi, Keller, Liang Shang, Johannes and others all predicted the olfactory perception from odor molecular structure and odor physicochemical characteristics, continued the research in 2007 and explored it at a deeper level [10]-[12]. These studies have shown that odor's pleasantness can be reflected in some useful component of the molecule, while the E-nose acts on the whole odor. In addition, in human recognition and E-nose measurement, odor information is realized by some form of associative memory, which is used to store and recall previously encountered odors [13]. The prediction of pleasantness by E-nose is more in line with human olfactory mechanism.

When using E-nose to predict odor's pleasantness, previous researchers designed a manual feature extractor to obtain the characteristics of odor information. Such an algorithm is not only designed to be large in workload, but also depends on the experience of the staff, and has poor versatility. But deep learning can be used to construct complex conceptual representation attribute categories or features by combining simple concepts to discover distributed feature representations of data. There is a stable effect on data learning without additional feature engineering requirements. At present, deep learning has been rapidly developed and widely used in computer vision, natural language processing, etc., but there are very few applications in terms of olfaction.

In this paper, we proposed a model for predicting odor's pleasantness with Convolutional Neural Network (POP-CNN). The contribution of this paper is reflected in three aspects.

1) This paper uses a convolutional neural network to process the E-nose response of odors. The dimensions of the odor samples conform to the dimensional requirements of the convolutional neural network.

2) In our model, the convolution kernels are designed to fit for the odor data. The kernels cover all the sensors, so them can catch the correlation mode of sensor response.

3) In order to reduce the dimension of odor data, we propose a non-uniform subsampling algorithm.

We proposed a new method for predicting pleasantness of odor, which get rid of the complex feature engineering while keeping the odor information as much as possible, and improves the learning efficiency.

The rest of this paper is organized as follows. Section Ⅱ reviews some related works. The introduction of Convolutional Neural Networks and the establishment of the POP-CNN Model are mentioned in Section Ⅲ. Section Ⅳ shows the detailed experimental process and results. Finally, the conclusion is given in Section Ⅴ.

## Ⅱ. RELATED WORKS

The olfactory pleasantness has become a hot topic in artificial olfactory. Many scholars have made outstanding contributions to this and laid a good foundation for follow-up research.

A.  Predicting pleasantness With physicochemical characteristics

In 2007, Rehan M. Khan et al. found that the main axis of perception odor is pleasantness using the molecular structure data of Dravnieks and the mature dimensionality reduction method of PCA. This work proved that the acquisition of odor pleasantness can be achieved by the physicochemical characteristics of odor [6].

In 2017, Hongyang Li, Bharat, Gilbert et al. integrated population and personal perception into a random forest model, effectively reducing the effects of noise and outliers, and accurately predicting individualization odor perception from large-scale chemical information [10]. Liang Shang et al. obtained the physicochemical parameters of odor molecules by using molecular calculation software (DRAGON), and extracted the characteristics of molecular parameters using PCA or Boruta algorithm as inputs to machine learning models (SVM, random forests and extreme learning machines) and compare their predictions [11]. American's Andreas Keller, Richard et al. used a regularized linear model and a random forest model to predict the odor perception in combination with the physicochemical information characteristics of odor molecules [12].

The methods for obtaining physicochemical characteristics of odors are generally two public data sets: Dravnieks and DRAGON, which are highly recognized worldwide. This method mainly has the following defects.

1) The real-time performance is low, some scenes need to detect pleasantness in real time, and the chemical formula of gas can't be known immediately.

2) The versatility is low, the gas is generally mixture of various substances in real life, not a single one. At this time, the method based on chemical characteristics cannot be solved well.

B.  Predicting pleasantness With E-nose

There are few studies on the measurement of odor signal data by E-nose sensors a for realizing the prediction of odor's pleasantness. In 2010, Rafi Haddad, Abebe Medhanie et al. used handcraft methods, such as the signal max value and latency to max, the time the signal reaches the half max and so on, to extract features from E-nose signal. In addition, 28 possible ratios of 8 MOX signals and 28 ratios of 8 QMB signals were added in each scent. Then, they input the features a single hidden-layer neural network with five hidden neurons to predicting the pleasant of odor [7]. Manual feature extractor has three main disadvantages.

1) The algorithm needs to be carefully designed, and the performance depends on the experience of the designer.

2) The versatility is relatively poor, and the characteristics suitable for certain odors may not be suitable for other odors.

3) The workload of the design algorithm is relatively large.

C.  Machine learning methods for E-nose

With the deepening of the research on E-nose technology, its application research has received people's attention and has been promoted and applied in the following fields.

In the food industry, the analysis of volatile components uses two conventional techniques: gas chromatography-mass spectrometry (GC-MS) and sensory expert analysis, but these two methods are time consuming and labor intensive and expensive, and the E-nose is a viable alternative. It can accurately classify black tea, identify different types of milk, whether the bread is moldy, classify beer and whether the quality of the tea is good [14]-[18]. The application of E-nose in food classification and quality testing is excellent.

In the flavor and fragrance industry, the E-nose is used for the certification of agarwood oils. Agarwood oils is a very precious fragrance tone, using E-nose and K-NN classifier to distinguish pure and mixed agarwood oils with an accuracy of 100% [19].

In medical diagnosis, one of the traditional methods is to extract some liquid from the human body for laboratory analysis, which is time consuming to operate. At present, many scholars have used E-nose to detect the odor exhaled by patients, and can diagnose and treat lung cancer, diabetes and kidney diseases simply and quickly [20]-[23].

Among the quality identification and classification of Chinese herbal medicines, most Chinese medicine practitioners regard the odor of Chinese herbal medicines as one of the important basis for identification of origin, variety and quality. The odor of the medicinal materials is related to the ingredients and properties involved. Each Chinese medicinal material has its own special odor, and some even have a strong pungent odor. After using the E-nose technology to obtain the odor information of Chinese herbal medicines, the method of machine learning can achieve the objective, accurate identification of Chinese herbal medicines for authenticity and quality assessment [24].

In people's daily life, there are some harmful odors such as $NH_3$, NO, CO, $NO_2$ and some flammable and explosive substances such as gasoline and fireworks in the surrounding environment or near the factory. These odors have a certain impact on human health, and there are hidden dangers. The E-nose can effectively monitor the harmful and toxic gases in our environment and keep them within a moderate range to protect our normal daily life [25]- [31].

All of the above work shows the importance of smell in people's lives. However, these studies only stayed on the characterization of the odor and did not give a perceptual description of the odor.

Ⅲ. The POP-CNN MODEL

A.  Brief Introduction to CNN

The most important feature of CNN is the convolution operation. It adopts an "end-to-end" learning method [32]. CNN can extract high-level features, so it has achieved good results in image applications. Such as image classification, image semantic segmentation, image retrieval, object detection and other machine vision problems.

The architecture of a typical CNN (Figure 1). It consists of two special types of layers—the convolutional layer and the pooling layer. The connection order is "Convolution-ReLU-(Pooling)" (the Pooling layer is sometimes omitted). These operational layers can be viewed as a complex function $f_{CNN}$ as a whole. The training of CNN is based on the "Loss" to update the model parameters and propagates the error back to the layers of the network. It can be understood as a direct "fitting" from the original data to the final goal.

The processing performed by the convolutional layer is a convolution operation. As shown in Fig. 2, the convolution operation is equivalent to "filter processing".

Convolution is the sum of two variables multiplied in a certain range. If the convolution variable is the sequence $x(n)$ and $h(n)$, the result of the convolution

$$y(n) = \sum_{-\infty}^{\infty} x(i)h(n-i) = x(n) * h(n) \tag{1}$$

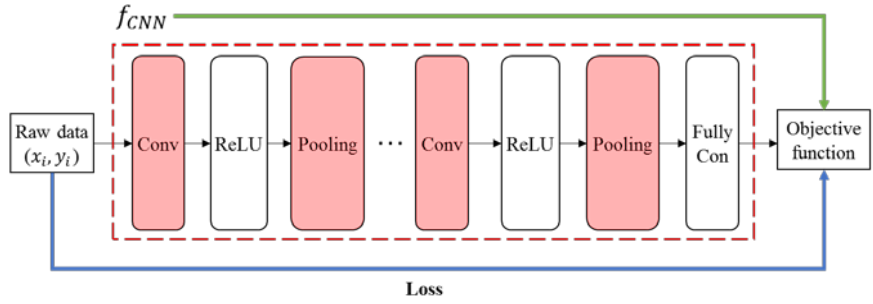

Fig. 1. The architecture of CNN.

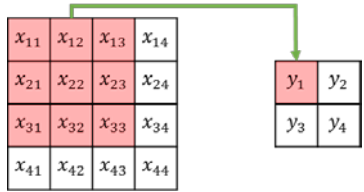

Fig. 2. Convolution operation. The size of the input data in the figure is (4, 4), the filter size is (3, 3), and the output size is (2, 2).

In the convolutional layer, the feature map of the previous layer and a set of convolution kernels called filters form the output features through the ReLU activation function. Generally, we have

$$X_j^l = f\left(\sum_{i \in M_j} X_i^{l-1} * K_{ij}^l + b_j^l\right) \tag{2}$$

Where $M_j$ represents a selection of the input maps, which is generally all- pairs or all-triples. The $b$ is the bias.

In the pooling layer, the input maps adopt the way of subsampling.

$$X_j^l = f(\beta_j^l down(X_j^{l-1}) + b_j^l) \tag{3}$$

Where down(·) represents the subsampling function. Each output map has its own multiplication bias β and an additional bias $b$ [33].

B. Non-uniform sampling algorithm

Generally, E-nose only more than 10 array sensors, and a few hundred seconds to measure odor's signal.

Assume that the number of sensors is $m$ and the acquisition time is $n$ seconds. Then the dimension of the odor data $(m * n)$ is very high, almost more than 1,000. Moreover, the number of odor samples is relatively small. So this is easily lead to make deep learning mode over-fitting.

As shown in Figure 3, the sensor's response goes through a rapid rise and fall period and then slowly drop.

According to the change rule of sensor's response, we propose a non-uniform data sampling algorithm, in which the sampling interval is small in the region where the data changes rapidly, and the sampling interval is large in the region where the data changes slowly.

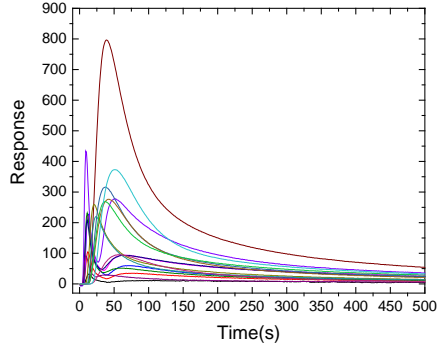

Fig. 3. The response curve of sensors. Each line represents a sensor.

This algorithm effectively reduces the dimension of odors data and avoids the problem of overfitting in deep learning. Moreover, the corresponding change mode of sensors is retained as much as possible, and the loss of information due to unreasonable sampling is avoided.

Suppose the first sensor responds with $r_{11}$ at $t_1$ second and $(t_1 + 1)$ with $r_{12}$. The response of the second sensor at $t_1$ second is $r_{21}$, $(t_1 + 1)$ is $r_{22}$, and the response of the $m$-th sensor at $t_1$ second is $r_{m1}$, $(t_1 + 1)$ is $r_{m2}$. Then within one second, the gradient of E-nose response is taken as the average of all sensors.

$$R_1 = \frac{(r_{12} - r_{11}) + \cdots (r_{m2} - r_{m1})}{m} \quad (4)$$

In order to be common to all odor samples, each sample is subjected the above calculation and the final average is taken to represent the gradient of the odors over time.

$$R_i = \frac{\sum_1^N R_j}{N} \quad (i = 1,2,\cdots,n) \quad (5)$$

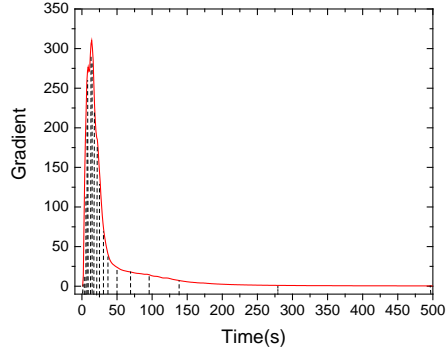

Fig. 4. The gradient of response. Red line is the gradient of odor samples at every second. Sampling the sensor response at the time corresponding to the dashed line (T=400).

The gradient of training samples is shown in fig. 4. From the first second, the summation operation of the gradient is performed.

$$S_i = R_i + R_{i+2} + \cdots + R_{i+k} \ (k < n) \qquad (6)$$

When the sum value exceeds a certain threshold ($S_i > T$), sampling is performed. Then the next sum is operated. And sample the last time in the last second (whether the sum value exceeds the threshold or not). The algorithm is summarized in **Algorithm 1**.

---

**Algorithm 1:** Non-uniform sampling algorithm

**Input:** Sensors response values of odors;
Parameter T, T is a threshold;
**Procedure:**
The odors are collected for $n$ seconds.
*Step 1:* Calculate the gradient for each sample according to formula (4), then find the average of all samples as the final gradient $R$ according to formula (5);
*Step 2:* From the first second, the cumulative gradient is added in order according to formula (6);
*Step 3:* Sampling once in the first second, and the next sampling when $S_i > T$. And sample the last time in the last second.

---

We use this algorithm to preprocess odor data.

C. Model for predicting odor's pleasantness by convolution neural network

The vector representation of each odor sample is very similar to the vector representation of the image (28*28 or 256*256, etc.). The E-nose uses array sensor technology, in which all sensors generate response signals during odor's collection, rather than a specific sensor. Therefore, the odor sample is composed of two dimensions: "number of sensors" and "response time series". Based on this, using CNN to realize the odor's pleasantness prediction with E-nose is reasonable and has great possibility of success. But the odor's data is special. When the E-nose collects odors, all sensors produce different degrees of response, and the resulting data format is a long-term sequence, which is much larger than the number of sensors. So we chose a structure different from the traditional CNN.

The architecture of the POP-CNN model is shown in Fig. 5. It consists of two convolutional layers and one fully connected layer.

The input layer of the training process is the odor's information collected by the E-nose after normalization. The input size is (238*1*16*250), it means there are 238 samples, each containing 1 channel and the size is (16*250). Let $x_i \epsilon R^h$ be the value of the $i$-th second of the $h$-dimensional E-nose sensor signal. The odor collected by $n$ sensors at the same time is expressed as

$$X_n = x_1 + x_2 + \cdots + x_h, h = 250 \tag{7}$$

$$X = X_1 + X_2 + \cdots + X_n, n = 16 \tag{8}$$

Where $+$ is just concatenation, not a summation. The filter is $k \in R^{nh}$. Let $x_{i:j}$ be the value from the $i$-th to the $j$-th second. A feature $c_i$ is generated after the filter convolution operation of $x_{i:j}$ by

$$c_i = f(x_{i:j} * k + b) \tag{9}$$

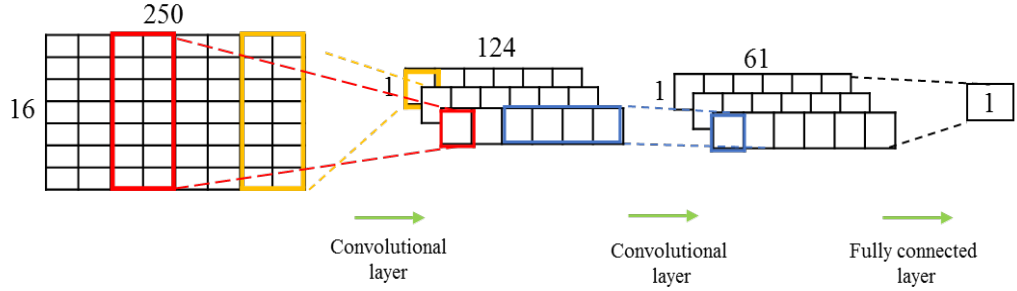

Fig.5. POP-CNN Model. There are two convolutional layers and one fully connected layer. The filter (16*4) is used in the first convolution to include all 16 sensors.

Where $b \epsilon R$ is the bias and $f(\cdot)$ is a non-linear function such as the ReLU. A convolution operation uses multiple filters, one filter corresponds to one feature, and multiple filters acquire multiple features to form a feature map.

$$C = [c_1, c_2, \cdots c_m] \tag{10}$$

The size of $m$ is related to the window size of the filter and the stride size of the convolution operation.

Since there are only 16 sensors, and each sensor collects odors for several hundred seconds. The width of response value matrix is larger than its height. In our work, the convolution kernel of the first filter operation is (16*4). This allows 16 sensors to be retained directly and let the filter do convolution only in the lateral direction of the data matrix.

When the odor is collected, the response of sensors rises or falls steadily, and the time is relatively long. In a short period of 1 or 2 seconds, these sensors do not change greatly. Instead of using the pooling layer, our POP-CNN chooses to increase the stride of the convolutional layer. By replacing the pooling layer with this method, the filter can avoid wasting time in some same windows and improving learning efficiency.

Several output feature maps are generated after the first convolution, and their size is (1*124). Then the second convolution operation is performed using (1*4) filter and the stride is the same as the first convolution operation. The pooling layer is omitted again. ReLU is applied to each convolutional layer. The last fully connected layer outputs a feature map as a predictor of odor pleasantness.

## IV. EXPERIMENTS

### A. Dataset

This paper uses the raw data of the E-nose and the score of human odor pleasantness used by Rafi. All sample data was visualized to determine whether the test set data can be predicted by the trained POP-CNN, as shown in Figure 6.

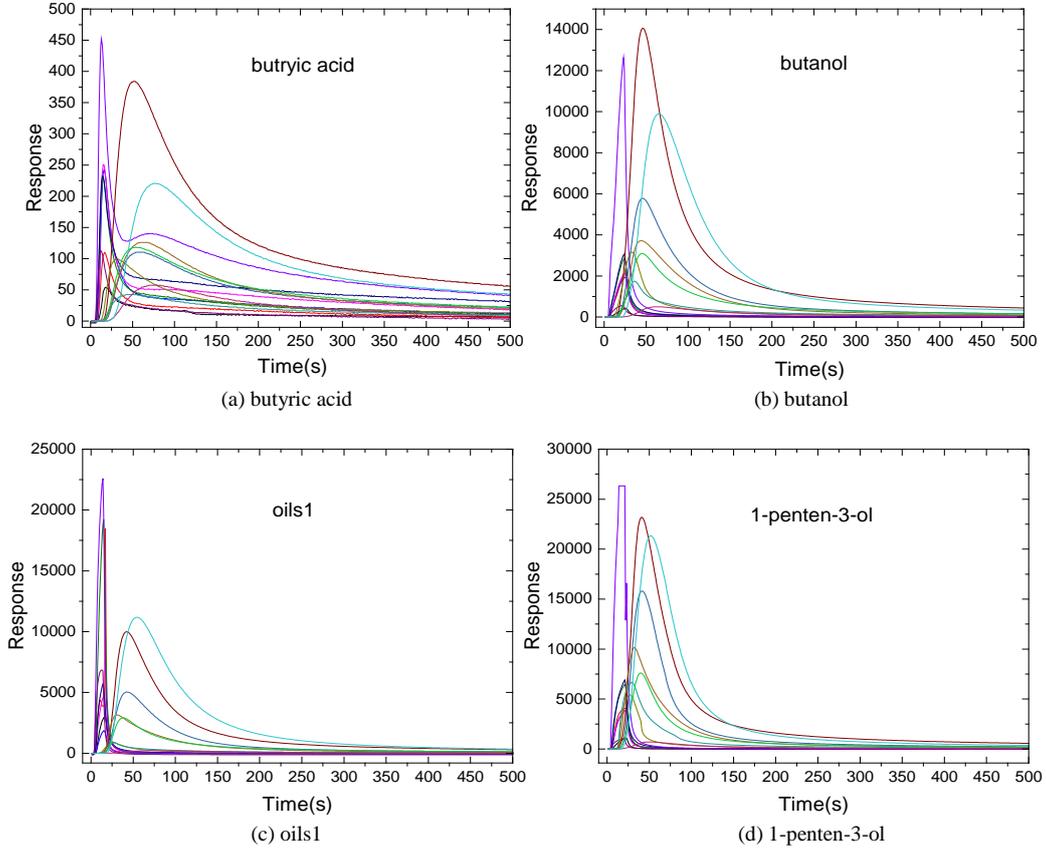

Fig.6. Data visualization. (a)(b) Two examples of training set visualization.
(c) One example of essential oils set visualization. (d) One example of novel odors set visualization.

These odors in his paper collected by the commercial MOSESII E-nose which contains eight metal-oxide (MOX) sensors and eight quartz microbalance (QBM) sensors [27], each of which responds to odors. In other words, according to people's usual concept, the response signal (whether voltage or resistance signal) generated by the sensor due to external stimulation is necessarily a one-dimensional vector related to the time series. Then the response of the E-nose consisting of array sensors to an odor is necessarily a time-series vector produced by all sensors. The MOSESII E-nose has a total of 16 sensors. Assuming that the odor is collected for $h$ seconds using an E-nose, each sample is a ($16*h$) dimensional vector, as shown in Table Ⅰ.

It can be seen from Fig. 6 that the second half of each sample data has almost no significance. We removed some redundant features by direct deletion and uniform sampling, reduced the sample dimension, and the processed sample dimensions is (16*250). In order to meet the requirements of the POP-CNN model for the sample input dimension, combined with the number of samples, we converted the E-nose raw data into (238*1*16*250), (108*1*16*250) and (95*1*16*250).

Table I

E-nose signal

| Sensor number | E-nose response | | | | | | |
|---|---|---|---|---|---|---|---|
| 1 | -18.667 | -16.667 | 88.333 | 481.33 | 975.33 | … | -16.667 |
| 2 | -3.6 | 14.4 | 527.4 | 1711.4 | 2870.4 | … | -3.6 |
| 3 | -2.2 | 5.8 | 119.8 | 355.8 | 631.8 | … | -2.2 |
| 4 | -20.2 | 82.8 | 684.8 | 1423.8 | 2095.8 | … | -21.2 |
| 5 | -71 | 282 | 2082 | 4828 | 7688 | … | -74 |
| 6 | -1.5 | 249.5 | 878.5 | 1920.5 | 3002.5 | … | -12.5 |
| 7 | -64.25 | 3591.8 | 7916.8 | 11439 | 14283 | … | -75.25 |
| 8 | -137 | 2270 | 3792 | 4918 | 5709 | … | -136 |
| 9 | 0 | 0 | 0 | 0 | 0 | … | 61 |
| 10 | -0.3 | -0.3 | -0.3 | -0.3 | -0.3 | … | -0.3 |
| 11 | 0.1 | 0.1 | 0.1 | 0.1 | 0.1 | … | 19.1 |
| 12 | -0.1 | -0.1 | -0.1 | -0.1 | -0.1 | … | 4.9 |
| 13 | 0.4 | -0.6 | 0.4 | -0.6 | 0.4 | … | 16.4 |
| 14 | 0 | 0 | 0 | 0 | 0 | … | 30 |
| 15 | 0 | 0 | 0 | 0 | 0 | … | 2 |
| 16 | 0 | 0 | 0 | 0 | 0 | … | 73 |

B.  Results and Discussion

In designing the experiment, in order to verify the stability of the POP-CNN model as much as possible, all the data was divided into a training set and two test sets. The specific method is shown in Figure 7.

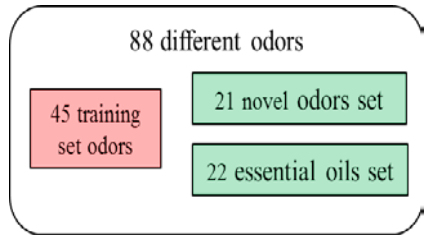

Fig. 7. Number of odor per data set. There are three data sets: the training set and the two test sets (untrained mixed essential oils set and novel odors set).

The training label for the POP-CNN model is the odor pleasantness score calculated using the visual analogue scale (VAS), which is, to be precise, the median of all volunteers participating in the experiment.

We trained our model using stochastic gradient descent. The number of samples per training was 14 and the momentum was 0.8. We found that a small amount of attenuation can make the learning of the POP model more stable. We use the same learning rate for all layers. Following the general rule of adjusting the learning rate, first set the learning rate to 0.01, then divide the learning rate by 10, and stop the learning rate change when the loss function is basically no longer improved. The final learning rate is 0.0001.

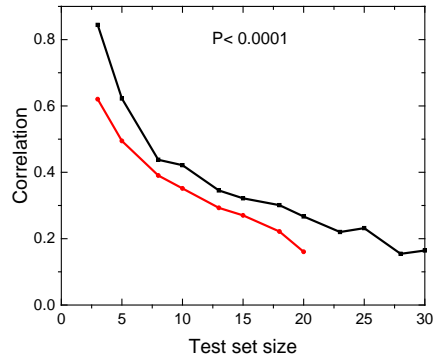

Fig. 8. Train-Validation experiment. Red line: The total number of odors used in this experiment is 45. And run the algorithm 20 times per experiment. Black line: The initial training set of 67 odors.

Using the training set-verification set scheme, the POP-CNN model was trained to predict the median of the scent pleasantness in the validation set. For the 45 odors of the original training set, they were randomly divided into training set and verification set. As the number of training sets increased, the correlation between the median value of the E-nose prediction and the human score gradually increased. For the 40 odor training sets, the correlation was 0.4947 ($P < 0.0001$; Figure 8).

Twenty-two essential oils data set were added to this experiment with an initial training set of 67 odors and a correlation of 0.4947 for the 40 odor training sets ($P < 0.0001$; Figure 8).

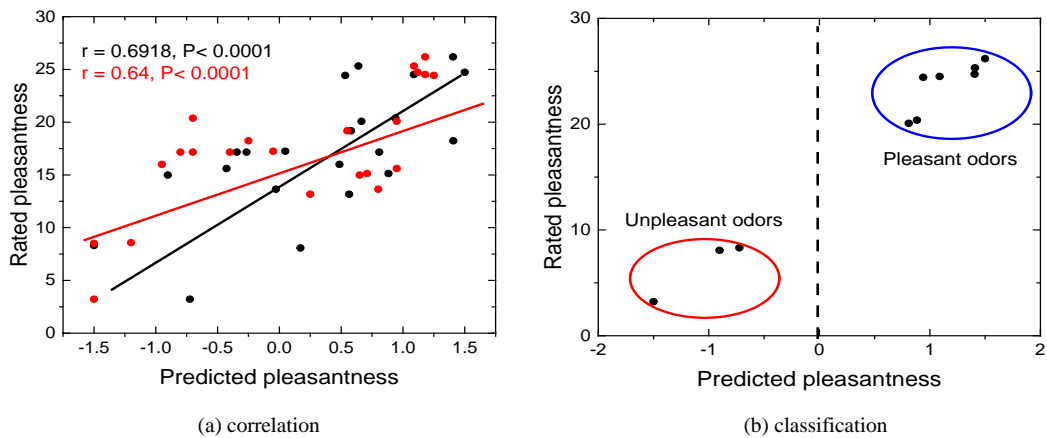

(a) correlation    (b) classification

Fig. 9. Predicting pleasantness of essential oils. (a) Correlation between the pleasantness of electronic nose prediction and human. (b) Pleasant and unpleasant odor classification.

The performance of the POP-CNN model was verified using odor data that was not trained to be used as a test set. Rafi used 22 essential oils odor mixtures made of unknown ingredients in his article, and we only modified it to fit the POP-CNN model. The POP-CNN model was trained using training sets of 45 odors and their corresponding human pleasantness scores. The median of the pleasantness scores of these odor mixtures was predicted using the already trained model, and the correlation between the E-nose prediction and the median human scores was $r = 0.6918$ (run 30 times, $P < 0.0001$; Figure 9(a)), while the correlation obtained in the Rafi article was $r = 0.64$. Then, the correlation between each person's pleasantness score and the human median was 0.72, so the machine-human correlation was 96% of the human-human correlation ($0.6918 / 0.72 * 100 = 96$).

To further verify the robustness of the POP-CNN model, an additional set of 21 novel odors was used. The volunteers who participated in the scoring of this group of odors did not participate in the training set. In this case, the correlation between the E-nose prediction and the median human score was r = 0.5070 (run 10 times, P < 0.0001; Figure 10(a)), while the correlation obtained in the Rafi article was r = 0.45. The correlation between each person's score and the human median was 0.55, and the machine-human correlation was 92% of the human-human correlation.

If the analysis is limited to absolute pleasantness and unpleasant odor, the odor of 10-20 pleasantness levels in the Human Pleasantness Rating Scale was removed. If the E-nose predicts a value above zero, then the odor is pleasant, otherwise it is unpleasant. Whether it is a blend of essential oils and novel odors set, the odors were divided into two groups with an accuracy of 99% (Figure 9(b) and Figure 10(b)).

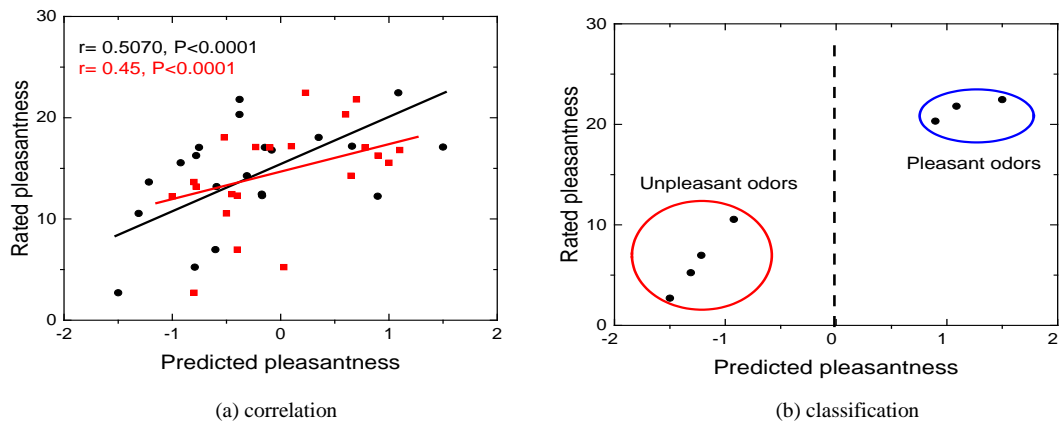

(a) correlation  (b) classification

Fig.10. Predicting pleasantness of novel odors. (a) Correlation between the pleasantness of electronic nose prediction and human. (b) Pleasant and unpleasant odor classification.

## V. CONCLUSION

Using the E-nose to obtain odor information is a good way to solve the problem of stimulus-perception. In this paper, according to the characteristics of odor information obtained by E-nose, we designed a model for predicting odor's pleasantness by CNN. And a simple CNN with a small number of layers performs very well. The correlation of our POP-CNN model is 8% higher than the Rafi in the essential oils set, and 10% higher in the novel odors set. This model can be used not only for pleasure prediction, but also for the classification and detection of odors, and to assess the taste of new odors in the perfume industry or environmental monitoring.